\documentclass{article}

\usepackage{microtype}
\usepackage{graphicx}
\usepackage{caption}
\usepackage{subcaption}
\usepackage{booktabs} 

\usepackage[english]{babel}
\usepackage[utf8]{inputenc}
\usepackage{amsmath}
\usepackage{amssymb}
\usepackage{graphicx}
\usepackage[colorinlistoftodos]{todonotes}
\usepackage{eurosym}
\usepackage{hyperref}
\usepackage{natbib}
\usepackage{mathtools}
\usepackage{bbm}
\usepackage[T1]{fontenc}
\usepackage{nicefrac}
\usepackage{bm}
\usepackage[thinc]{esdiff}
\usepackage{listings}
\usepackage{mathrsfs}

\usepackage{amsthm}
\newtheorem{theorem}{Theorem}

\DeclareMathOperator{\leakyrelu}{LeakyReLU}
\DeclareMathOperator{\softmax}{softmax}

\makeatletter
\renewcommand*\env@matrix[1][*\c@MaxMatrixCols c]{%
  \hskip -\arraycolsep
  \let\@ifnextchar\new@ifnextchar
  \array{#1}}
\makeatother

\makeatletter
\newsavebox{\@brx}
\newcommand{\llangle}[1][]{\savebox{\@brx}{\(\m@th{#1\langle}\)}%
  \mathopen{\copy\@brx\kern-0.5\wd\@brx\usebox{\@brx}}}
\newcommand{\rrangle}[1][]{\savebox{\@brx}{\(\m@th{#1\rangle}\)}%
  \mathclose{\copy\@brx\kern-0.5\wd\@brx\usebox{\@brx}}}
\makeatother

\renewcommand{\vec}[1]{\mathbf{#1}}%

\usepackage{hyperref}

\usepackage[accepted]{icml2021}

\icmltitlerunning{GRAND: Graph Neural Diffusion}

\begin{document}

\twocolumn[
\icmltitle{GRAND: Graph Neural Diffusion}



\icmlsetsymbol{equal}{*}

\begin{icmlauthorlist}
\icmlauthor{Benjamin P. Chamberlain}{equal,twit}
\icmlauthor{James Rowbottom}{equal,twit}
\icmlauthor{Maria Gorinova}{twit}
\icmlauthor{Stefan Webb}{twit}
\icmlauthor{Emanuele Rossi}{twit}
\icmlauthor{Michael M. Bronstein}{twit,imp,idsia}
\end{icmlauthorlist}

\icmlaffiliation{twit}{Twitter Inc., London, UK}
\icmlaffiliation{imp}{Imperial College London, UK}
\icmlaffiliation{idsia}{IDSIA/USI, Switzerland}
\icmlcorrespondingauthor{Ben Chamberlain}{bchamberlain@twitter.com}

\icmlkeywords{Graph Neural Networks, Partial Differential Equations, Diffusion Processes}

\vskip 0.3in
]



\printAffiliationsAndNotice{\icmlEqualContribution} 

\begin{abstract}
We present Graph Neural Diffusion (GRAND) that approaches deep learning on graphs as a continuous diffusion process and treats Graph Neural Networks (GNNs) as discretisations of an underlying PDE. In our model, the layer structure and topology correspond to the discretisation choices of temporal and spatial operators. Our approach allows a principled development of a broad new class of GNNs that are able to address the common plights of graph learning models such as depth, oversmoothing, and bottlenecks. Key to the success of our models are stability with respect to perturbations in the data and this is addressed for both implicit and explicit discretisation schemes. We develop linear and nonlinear versions of GRAND,  which achieve competitive results on many standard graph benchmarks.
\end{abstract}

\section{Introduction}

Machine learning on graphs and graph neural networks (GNNs) have been shown to be successful in a broad range of problems across different domains,  %
extending way beyond machine learning. Important results have been achieved in the physical sciences \cite{Li2020c,Li2020d}, where \textit{partial differential equations} (PDEs) have traditionally been the dominant modelling paradigm.

GNNs are in fact intimately connected to differential equations. The seminal work of  \citet{Scarselli2008} was concerned with finding the fixed points of differential equations using the Almeida-Pineda algorithm~\cite{almeida1987learning, pineda1987generalization}. The currently predominant message passing paradigm~\cite{Gilmer2017} can be modelled as a differential equation. More recently, diffusion processes  have been shown to be an effective preprocessing step for graph learning~\cite{Klicpera2019}.

PDEs are among the most studied mathematical constructions, with a vast literature 
dating back at least to Leonhard Euler in the eighteenth century. This includes various  discretisation schemes, numerical methods for  approximate solutions, and theorems for their existence and stability. 
Historically, PDE-based methods have been used extensively in signal and image processing \cite{perona1990scale}, computer graphics \cite{sun2009concise}, and more recently, in machine learning \cite{Chen2018a}.

Our goal is to show that the tools of PDEs can be used to understand existing GNN architectures and as a principled way to develop a broad class of new methods.
%
%
We focus on GNN architectures that can be interpreted  as information diffusion on graphs, modelled by the diffusion equation. In doing so, we show that many popular GNN architectures can be derived from a single mathematical framework by different choices of the form of diffusion equation and discretisation schemes. 
%
%
Standard GNNs are equivalent to the explicit single-step Euler scheme that is inefficient and requires small step sizes. 
We show that more advanced, adaptive multi-step schemes such as Runge-Kutta perform significantly better and using implicit schemes, which are unconditionally stable, amounts to larger multi-hop diffusion operators. 
%
%
Choosing different spatial discretisation amounts to {\em graph rewiring}, a technique recently used to improve the performance of GNNs ~\cite{Klicpera2019,Alon2020}. 
We show that appropriate choices within our framework allow the design of deep GNN architectures with tens of layers. This is a feat hard to achieve otherwise due to feature oversmoothing \cite{NT2019, Oono2019} and bottlenecks~\cite{Alon2020} -- phenomena that are recognised as a common plight of most graph learning architectures. 

\paragraph{Main contributions}
We describe a broad new class of GNNs  based on the discretised diffusion PDE on graphs and study different numerical schemes for their solution.  
Second, we provide stability conditions for these schemes. 
Finally, based on our model, we develop linear and nonlinear 
Graph Neural Diffusion (GRAND) architectures 
that perform competitively on many popular benchmark datasets. 
We show detailed ablation studies shedding light on the choice of numerical schemes and parameters. 

\section{Background} \label{sec:background}

Central to our work is the notion of diffusion processes. 
In this section, we provide a concise background on diffusion equations in the continuous setting, on which we build in Section~\ref{sec:graph_diffusion}  to develop similar notions on graphs. As we are concerned with continuous analogues of graph diffusion and graphs are associated with a broad array of underlying geometries, it is inadequate to formulate these processes in simple flat spaces and more general Riemannian manifolds are required.

\paragraph*{Diffusion equation}
We are interested in studying diffusion processes on $\Omega$. 
Informally, diffusion describes the movement of a substance from regions of higher to lower concentration. For example, when a hot object is placed on a cold surface, heat will diffuse from the object to the surface until both are of equal temperature.

Let $x(t)$ denote a family of scalar-valued functions on $\Omega \times [0,\infty)$ representing the distribution of some property (which we will assume to be temperature for simplicity) on $\Omega$ at some time, and let $x(u,t)$ be its value at point $u\in \Omega$ at time $t$.  
According to Fourier's law of heat conduction,  the heat flux  
%
\begin{align}
    h = -g \nabla x, \notag
\end{align}
is proportional to the temperature {\em gradient} $\nabla x$, where 
%
$g$ is the {\em diffusivity} describing the thermal conductance properties of $\Omega$. 
%
An idealized {\em homogeneous} setting assumes that $g$ is a constant scalar throughout $\Omega$. More generally, the diffusivity is a {\em inhomogeneous} (position-dependent) function that can be scalar-valued (in which case it simply scales the temperature gradient and is {\em isotropic}) or matrix-valued (in which case the diffusion is said to be {\em anisotropic}, or direction-dependent). 
%
%
The continuity condition $x_t = -\mathrm{div}(h)$ (roughly meaning that the only change in the temperature is due to the heat flux, as measured by the {\em divergence} operator, i.e., heat is not created or destroyed),  
leads to a PDE referred to as the  {\em (heat) diffusion equation},\vspace{-1mm} 
$$
\frac{\partial x(u,t)}{\partial t} = \mathrm{div} [ g(u,x(u,t),t) \nabla x(u,t) ],
$$
%
with the {initial condition} $x(u,0) = x_0(t)$; for simplicity, we assume no boundary conditions. 
%
The choice of the diffusivity function determines if the diffusion is 
homogeneous ($g=c$), inhomogeneous ($g(u,t)$), or anisotropic ($A(u,t)$). 
%
%
In the isotropic case, the diffusion equation can be expressed as $\frac{\partial x(u,t)}{\partial t} = \mathrm{div}( c\nabla x) = c\Delta x$, where $\Delta x = \mathrm{div}(\nabla x)$ is the {\em Laplacian} operator. 
%

\paragraph*{Diffusion on manifolds}
In our discussion so far we assumed some abstract domain $\Omega$. The structure of the domain is manifested in the definition of the spatial differential operators in the diffusion PDE. 
In a general setting, we model $\Omega$ as a Riemannian manifold, and let $\mathcal{X}(\Omega)$ and $\mathcal{X}(T\Omega)$ denote the spaces of {\em scalar} and {\em (tangent) vector fields} on it, respectively. 
%
%
We denote by 
$\langle x, y\rangle$ 
and $\llangle \mathscr{X}, \mathscr{Y}\rrangle$ 
the respective inner products on $\mathcal{X}(\Omega)$ and $\mathcal{X}(T\Omega)$. %
Furthermore, we denote by $\nabla: \mathcal{X}(\Omega) \rightarrow \mathcal{X}(T\Omega)$ and $\mathrm{div} = \nabla^*: \mathcal{X}(T\Omega) \rightarrow \mathcal{X}(\Omega)$ the {\em gradient} and {\em divergence} operators, which are adjoint w.r.t. the above inner products:
$
\llangle \nabla x, \mathscr{X} \rrangle = \langle x, \mathrm{div}(\mathscr{X}) \rangle
$.
Informally, the gradient $\nabla x$ of a scalar field $x$ is a vector field providing at each point $u\in \Omega$ the direction $\nabla x(u)$ of the steepest change of $x$.
The divergence $\mathrm{div}(\mathscr{X})$ of a vector field $\mathscr{X}$ is a scalar field providing, at each point, the flow of $\mathscr{X}$ through an infinitesimal volume. 
The Laplacian $\Delta x$ can be interpreted as the local difference between the value of a scalar field $x$ at a point and its infinitesimal neighbourhood.

\paragraph*{Applications of diffusion equations}
In image processing, diffusion equations were used for nonlinear filtering of images. Given an image $x$ defined on $\Omega = [0,1]^2$, the non-homogeneous isotropic diffusion equation \vspace{-1mm} 
$$
\frac{\partial x(t)}{\partial t} = \mathrm{div}\left[g(\| \nabla x(u,t)\|) \nabla x(u,t) \right],\vspace{-1mm}
$$
applied to the input image $x(u,0) = x_0(u)$ as the initial condition, 
is often referred to as {\em Perona-Malik diffusion} or (erroneously) anisotropic diffusion~~\cite{perona1990scale}. The scalar function $g \propto \| \nabla x(u,t)\|^{-1}$ is referred as an {\em edge indicator} and is designed to prevent diffusion across discontinuities (edges) in the image, thus preserving its sharpness while at the same time removing the noise. 
%
%
In computer graphics and geometry processing, non-Euclidean diffusion equations were studied as shape descriptors. 

\section{Diffusion equations on graphs} \label{sec:graph_diffusion}

%

We now define \textit{diffusion equations on graphs}, analogous to Section~\ref{sec:background} and argue that formalizing GNNs under the diffusion equation framework provides a principled and rigorous way to develop new architectures for graph learning.

\subsection{Graph diffusion equation}
Let $\mathcal{G}=(\mathcal{V},\mathcal{E})$ be an undirected graph with $|\mathcal{V}|=n$ nodes and $|\mathcal{E}|=e$ edges, and let $\mathbf{x}$ and $\mathbf{\mathscr{X}}$ denote features defined on nodes and edges respectively.\footnote{For simplicity, we assume these features to be scalar-valued and refer to them as node and edge fields, by analogy to scalar and vector fields on manifolds. In the rest of the paper, we assume vector-valued node features, a straightforward extension. }
The node and edge fields can be represented as $n$- and $e$-dimensional vectors assuming some arbitrary ordering of nodes. 
We adopt the same notation for the respective inner products:\vspace{-4pt}
\begin{eqnarray*}
\langle \mathbf{x}, \mathbf{y}\rangle = \sum_{i \in \mathcal{V}} x_i y_i \quad\quad\quad
\llangle \mathbf{\mathscr{X}}, \mathbf{\mathscr{Y}}\rrangle = 
\sum_{i>j} w_{ij} \mathscr{X}_{ij} \mathscr{Y}_{ij}\vspace{-8pt}
\end{eqnarray*}
Here, $w_{ij}$ denotes the {\em adjacency} of $\mathcal{G}$: $w_{ij} = w_{ji} = 1$ iff $(i,j) \in \mathcal{E}$. We tacitly assume edge fields to be {\em alternating}, so $\mathscr{X}_{ji} = - \mathscr{X}_{ij}$, and no self-edges, so $(i,i) \notin \mathcal{E}$. %
The gradient $(\nabla \mathbf{x})_{ij} = x_j - x_i$ assigns the edge $(i,j) \in \mathcal{E}$ the difference of its endpoint features and is alternating by definition. 
Similarly, the divergence $(\mathrm{div}(\mathbf{\mathscr{X}}))_i$ assigns the node $i$ the sum of the features of all edges it shares:\vspace{-1mm}
$$
(\mathrm{div}(\mathbf{\mathscr{X}}))_i = \sum_{j : (i,j) \in \mathcal{E}} \mathscr{X}_{ij} = \sum_{j=1}^n w_{ij} \mathscr{X}_{ij}\vspace{-1mm}
$$
The two operators are adjoint, 
$\llangle  \nabla \mathbf{x}, \mathbf{\mathscr{X}}\rrangle = \langle  \mathbf{x}, \mathrm{div}(\mathbf{\mathscr{X}}) \rangle$. 
%

We consider the following diffusion equation on the graph \vspace{-1mm}
\begin{align}
\frac{\partial \mathbf{x}(t)}{\partial t} = \mathrm{div}[ \mathbf{G}(\mathbf{x}(t),t) \nabla \mathbf{x}(t) ]\vspace{-1mm}
\label{eq:graph_diffusion}
\end{align}
with an initial condition $\mathbf{x}(0)$. Here we denote by $\mathbf{G} = \mathrm{diag}( a(x_i(t),x_j(t),t) )$ an $e\times e$ diagonal matrix and $a$ is some function determining the similarity between nodes $i$ and $j$. While in general $a(x_i,x_j,t)$ can be time-dependent, we will assume $a = a(x_i, x_j)$ for the sake of simplicity. 
%
Plugging in the expressions of $\nabla$ and $\mathrm{div}$, we get
\begin{equation}
\frac{\partial}{\partial t}\mathbf{x}(t) = (\mathbf{A}(\mathbf{x}(t))-\mathbf{I})\mathbf{x}(t) = \bar{\mathbf{A}}(\mathbf{x}(t))\mathbf{x}(t)
\label{eq:ode}
\end{equation}
where $\mathbf{A}(\mathbf{x}) = (a(x_i,x_j))$ is the $n\times n$ {\em attention matrix} with the same structure as the adjacency of the graph (we assume $a_{ij}=0$ if $(i,j) \notin \mathcal{E}$). 
Note that in the setting when $\mathbf{A}(\vec{x}(t)) = \mathbf{A}$ we get a linear diffusion equation that can be solved analytically as $\vec{x}(t) = e^{\bar{\mathbf{A}}t} \vec{x}(0)$.

\subsection{Properties of the graph diffusion equation}

Differential equation stability is closely related to the concept of robustness in machine learning; changes in model outputs should be small under small changes in inputs. 
Formally, a solution $\mathbf{x}(t)$ of the PDE is said to be stable, if given any $\epsilon>0$ there exists $\delta>0$ such that for any solution $\hat{\mathbf{x}}(t)$, such that $|\mathbf{x}(0)-\hat{\mathbf{x}}(0)| \leq \delta$, it is also the case that $|\mathbf{x}(t)-\hat{\mathbf{x}}(t)| \leq \epsilon$ for all $t \geq 0$.

In the linear case, it is sufficient to show that the eigenvalues of $\bar{\mathbf{A}}$ are non-positive 
(see Appendix~\ref{ap:stability} for proof)
%
For the general nonlinear case, we show 
\begin{align}
    \max_ix_i(0) \geq x_i(t) \geq \min_ix_i(0) \quad \forall t\geq 0,
    \label{eq:max_min_bounds}
\end{align}
which follows from (i) the function $\bar{A}(\vec{x})\vec{x}$ being continuous in $\vec{x}$, (ii) the largest component of $\vec{x}(t)$ not increasing in time, and (iii) the smallest component is not decreasing in time. 

Condition (i) holds as $\bar{\mathbf{A}}$ is a composition of Lipschitz-continuous functions (cf. equation~\eqref{eq:attention}). Defining indices $k = \arg\max_i x_i$ and $l = \arg\min_i x_i$ we have 
\begin{align}
    \frac{\partial x_k}{\partial t} &= \sum_{j}\bar{a}_{kj}(x)x_j \leq x_k \sum_j \bar{a}_{kj} = 0 \\
    \frac{\partial x_l}{\partial t} &= \sum_{j}\bar{a}_{lj}(x)x_j \geq x_l \sum_j \bar{a}_{lj} = 0 
    \label{eq:k_inequality}
\end{align}
since $\mathbf{A}$ is right stochastic, which proves (ii) and (iii).

Furthermore, the derivative $\frac{\partial}{\partial x} \mathbf{A}(\mathbf{x})$ is Lipschitz-continuous (from the definition of the attention function we use), 
Taken together with continuity in time, the requirements of Picard-Lindel{\"o}f are satisfied and our PDE is also well posed.





\subsection{Solving the graph diffusion equation}


There are a wide range of numerical techniques for solving nonlinear diffusion equations. Our method most resembles the Method of Lines (MOL) where a finite difference method discretises the spatial derivatives, leaving a linear system of ODEs on the temporal axis that can be solved with numerical integrators. 
On a graph, the spatial operators are already discrete and follow the structure of the input graph; nevertheless, we show that different structures can be used, thus decoupling the input and computational graph. 

\begin{figure}
    \centering\vspace{-4mm}
    \includegraphics[width=1.05\columnwidth]{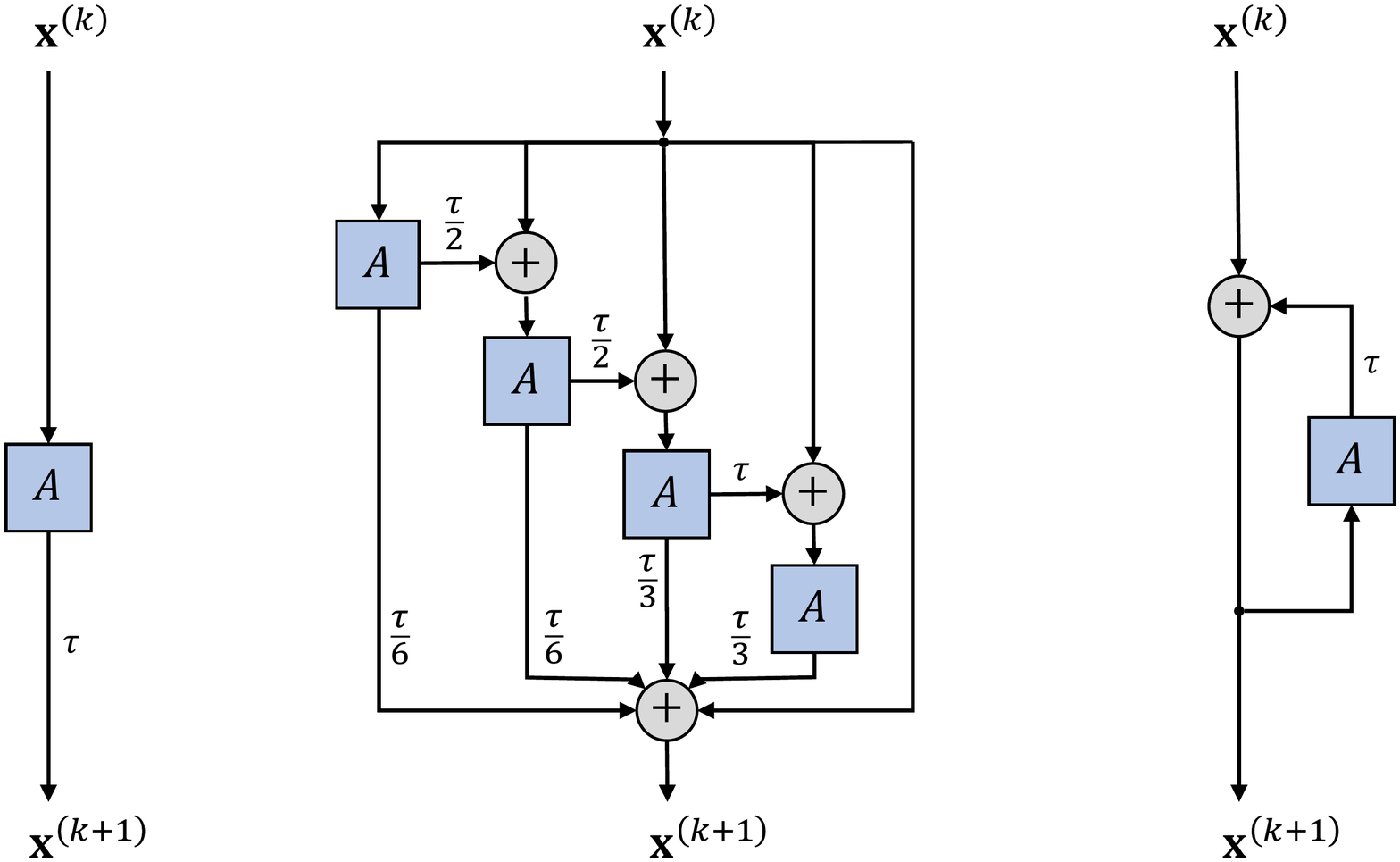}\vspace{-7mm}
    \caption{
    Block diagrams of (left to right) explicit Euler, 4th order Runge-Kutta, and implicit Euler schemes.\vspace{-3mm}
    }
    \label{fig:rk4_step}
\end{figure}

For temporal discretisation, there exist two main schemes: \emph{explicit} and \emph{implicit}. Furthermore, we can distinguish between {\em single-step} and {\em multi-step} schemes; the latter use multiple function evaluations at different times to compute the next iterate (see Figure~\ref{fig:rk4_step}).

\paragraph{Explicit schemes.}
The simplest way to  discretise Equation~\eqref{eq:graph_diffusion} is using the forward time difference: \vspace{-1mm}
\begin{equation}
\frac{x^{(k+1)}_ i - x^{(k)}_i}{\tau} = \sum_{j: (i,j) \in \mathcal{E}}
\hspace{-3pt} 
a\left(x^{(k)}_i, x^{(k)}_j\right) (x^{(k)}_j - x^{(k)}_i), \vspace{-1mm}
\label{eq:g_diff}
\end{equation}
where $k$ denotes the discrete time index (iteration), $\tau$ is the time step (discretisation parameter), and 
$a$ is assumed to be normalised, $ \sum_{j}  a(x^{(k)}_i, x^{(k)}_j) = 1$. 
%
%
%
Rewriting compactly in matrix-vector form,
$\frac{\mathbf{x}^{(k+1)} - \mathbf{x}^{(k)}}{\tau}= \left(\mathbf{A}(\mathbf{x}^{(k)})-I\right) \mathbf{x}^{(k)} = \mathbf{\bar{A}}(\mathbf{x}^{(k)}) \mathbf{x}^{(k)},$
leads to the {\em explicit} or {\em forward Euler scheme} (Figure~\ref{fig:rk4_step}, left):
\begin{equation}
\mathbf{x}^{(k+1)} = \left(\mathbf{I} + \tau \mathbf{\bar{A}}(\mathbf{x}^{(k)})\right) \mathbf{x}^{(k)}  
= \mathbf{Q}^{(k)} \mathbf{x}^{(k)},
\label{eq:g_diff_m}
\end{equation}
where for $a_{ij}^{(k)} = a(x_i^{(k)},x_j^{(k)})$, the matrix $\mathbf{Q}^{(k)}$ is given by 
$q_{ii}^{(k)} = 1-\tau \hspace{-5pt}\displaystyle\sum_{\ell: (i,\ell)\in \mathcal{E}}\hspace{-5pt} a_{i\ell}^{(k)}$, $q_{ij}^{(k)} = \tau a_{ij}^{(k)}$ if $(i,j) \in \mathcal{E}$, and $q_{ij}^{(k)} = 0$ otherwise.
%
This scheme is called explicit because the update $\mathbf{x}^{(k+1)}$ is deduced from $\mathbf{x}^{(k)}$ directly by the application of the diffusion operator $\mathbf{Q}^{(k)}$. 
The solution to the diffusion equation is computed by applying the scheme~(\ref{eq:g_diff_m}) multiple times in sequence, starting from some initial 
$\mathbf{x}^{(0)}$. 

\paragraph{Implicit schemes}
use a backward time difference, 
$\frac{\mathbf{x}^{(k+1)} - \mathbf{x}^{(k)}}{\tau}=  \mathbf{\bar{A}}\left(\mathbf{x}^{(k)}\right) \mathbf{x}^{(k+1)}  
$,
which leads to the (semi-){\em implicit} scheme (Figure~\ref{fig:rk4_step}, right):
\begin{equation}
\left(\mathbf{I} - \tau \bar{A}(\mathbf{x}^{(k)})\right) \mathbf{x}^{(k+1)}  
= \mathbf{B}(\mathbf{x}^{(k)}) \mathbf{x}^{(k+1)} = \mathbf{x}^{(k)}
\label{eq:g_diff_semimp}
\end{equation}
This scheme is called (semi-)implicit because it requires solving a linear system in order to compute the update $\mathbf{x}^{(k+1)}$ from $\mathbf{x}^{(k)}$, amounting to the inversion of $\mathbf{B}$. The efficiency of this step is crucially dependent on the structure of $\mathbf{B}$ --- for example, on grids, this matrix has a multi-diagonal structure, allowing $\mathcal{O}(n)$ inversion that was heavily exploited in PDE-based image processing applications \cite{Weickert1997}. 
In general, exact inversion is replaced with a few iterations of a linear solver.

\paragraph{Stability}
There exists a tradeoff between the number of iterations of the scheme $K$ and the time step size $\tau$. 
At the same time, the step size $\tau$ must be chosen in a way that guarantees that the scheme is stable. 
%
We summarise the stability results in the following theorems and provide additional details and proofs in Appendix~\ref{ap:stability}.

\begin{theorem}
\label{theo:stability_exp}
The explicit scheme~(\ref{eq:g_diff_m}) is stable for $0 < \tau < 1$. 
\end{theorem}

\begin{theorem}
\label{theo:stability_imp}
The implicit scheme~(\ref{eq:g_diff_semimp}) is unconditionally stable for any $\tau > 0$. 
\end{theorem}

\paragraph{Multi-step schemes}
use intermediate fractional time steps to obtain a higher-order numerical approximation, reusing the calculations for efficiency.  
Runge-Kutta (Figure~\ref{fig:rk4_step}, center) is among the most common multi-step schemes.
General linear multi-step methods 
calculate the subsequent iterate 
using a linear combination of previous iterates of the form, 
\begin{align}
    \sum^s_{j=0}\alpha_j \mathbf{x}^{(k+j)} &= \tau\sum^s_{j=0}\beta_j \bar{\mathbf{A}} (\mathbf{x}^{(k+j)}) \mathbf{x}^{(k+j)}.
\end{align}
and can be explicit or implicit depending $s$ and $\{ \alpha_j, \beta_j\}$. 

The (explicit) Adams--Bashford and (implicit) Adams--Moulton methods are classes of linear multi-step methods that set $\alpha_{s-1}=-1$ and $\alpha_{s-2}=\ldots=\alpha_0=0$. For both, the $\{\beta_j\}$ coefficients are solved for by interpolating the dynamics function at the points of the previous solutions, $\mathbf{x}^{(k+j)}$ with a polynomial of order highest order possible using the Lagrange formula and substituting this into the integral form of the ODE. The methods differ in that Adams--Moulton interpolates through $\mathbf{x}^{(k+s)}$ and is consequently implicit whereas the Adams--Bashford methods do not.
For Adams--Moulton methods, the implicit equations can be solved by Newton's method. 
Alternatively, one can use the predictor-corrector algorithm, which in this case takes an initial step with the explicit Adams--Bash method then multiple steps of Adams--Moulton, replacing the unknown $\mathbf{x}^{(k+s)}$ with the solution from the previous iteration, repeating until the difference between adjacent solutions is less than some threshold.
In our experiments, we use fourth-order methods, $s=4$. Additional details of multi-step schemes are provided in Appendix~\ref{sec:multistep}.

\paragraph{Adaptive step size}

Adaptive step size solvers estimate the error in 
each iteration, which is 
then compared to an error tolerance; the step size is adapted to either increase or reduce the error. The error is estimated by comparing two methods, one with order $p$ and one with order $p-1$. They are interwoven, i.e., they have common intermediate steps. As a result, estimating the error has little or negligible computational cost compared to a step with the higher-order method. Further details are given in Appendix~\ref{sec:multistep}.

%
%
%
%

\subsection{Connection to existing architectures}

Many GNN architectures can be formalised as a discretisation scheme of~(\ref{eq:graph_diffusion}). 
The discrete time index $k$ corresponds to a (convolutional) layer of the graph neural network. Running the diffusion for multiple iterations thus amounts to applying a GNN layer multiple times. 
%
In the diffusion formalism, the time parameter $t$ acts as a continuous analogy of the layers, in the spirit of Neural ODEs~\cite{Chen2018a}. This interpretation allows us to exploit more efficient numerical schemes and analyze the stability and convergence of the diffusion process.

The vast majority of GNN architectures are explicit single-step schemes of the form~(\ref{eq:g_diff_m}). %
For example, Equation~(\ref{eq:g_diff}) corresponds to the update formula of GAT \cite{Velickovic2018} with residual connection, assuming $a$ is a learnable attention function and no non-linearity is used between the layers. 
Our choice of a time-independent attention function
in the experiments in this paper amounts to all the layers {\em sharing the same parameters}. We will show that this is actually an advantage, as our models will be significantly more lightweight and less prone to overfitting.

The diffusion equation is a PDE, with temporal and spatial components. In the graph setting, the former is continuous while the latter is discrete. 
Thus, the diffusion operator $\mathbf{Q}$ inherits the structure of the adjacency of the input graph. 
However, it is possible to consider the graph as a discretisation of a continuous object and thus regard the graph diffusion operator as a discrete derivative. 
In the same way that different discretisations of continuous derivatives with different support can be chosen, we can {\em rewire} the graph and make the structure of $\mathbf{Q}$ different from the input one and possibly {\em learnable}.  
Multiple GNN architecture {\em de facto} use a different computational graph from the input one, whether for reasons of scalability (e.g. sampling used in GraphSAGE~\cite{Hamilton2017}), denoising the input graph~\cite{Klicpera2019}, or avoiding bottlenecks~\cite{Alon2020}.  
We argue that additional reasons are numerical convenience, to produce diffusion operators that are e.g. friendlier for matrix inversion.

In the following, we also show that the use of more efficient multi-step explicit schemes as well as unconditionally stable implicit schemes offers significant performance advantages.  
In particular, implicit schemes of the form~(\ref{eq:g_diff_semimp}) can be interpreted as multi-hop diffusion operators, since the inverse of $\mathbf{B}$ is typically dense (unlike $\mathbf{Q}$ in the explicit scheme~(\ref{eq:g_diff_m}) that has the same sparsity structure of the 1-hop adjacency matrix of the graph). 
%


\section{Graph Neural Diffusion}
We now describe Graph Neural Diffusion (GRAND), a new class of GNN architectures derived from the graph diffusion formalism.  
We assume a given graph $\mathcal{G}=(\mathcal{V},\mathcal{E})$ with $n$ nodes and $d$-dimensional node-wise features represented as a matrix $\mathbf{X}_{\mathrm{in}}$. 
%
GRAND architectures implement the learnable encoder/decoder functions $\phi$, $\psi$  and 
a learnable graph diffusion process, 
to produce node embeddings $\mathbf{Y} = \psi(\mathbf{X}(T))$, \vspace{-1mm}
\begin{align}
    \mathbf{X}(T) = \mathbf{X}(0) + \int_0^T \frac{\partial\mathbf{X}(t)}{\partial t} dt, \quad \quad
        \mathbf{X}(0) &= \phi(\mathbf{X}_{\mathrm{in}}) \notag 
\vspace{-6mm}
\end{align}
%

$\frac{\partial\mathbf{X}(t)}{\partial t} $ is given by the graph diffusion equation~\eqref{eq:graph_diffusion}. 
Different GRAND architectures amount to the choice of the learnable diffusivity function $\mathbf{G}$ and spatial/temporal discretisations of equation~\eqref{eq:graph_diffusion}.  
%

The diffusivity is modelled with an attention function $a(.,.)$. Empirically, scaled dot product attention~\cite{Vaswani2017} outperforms the \citet{Bahdanau2014} attention used in GAT~\cite{Velickovic2018}. The scaled dot product attention is given by 
\begin{align}
    a(\vec{X}_i,\vec{X}_j) = \softmax \left(\frac{(\mathbf{W}_K \vec{X}_{i} )^\top \mathbf{W}_Q \vec{X}_{j}}{d_k}\right),
    \label{eq:attention}
\end{align}
where $\mathbf{W}_K$ and $\mathbf{W}_Q$ are learned matrices, and $d_k$ is a hyperparameter determining the dimension of $W_k$.  
We use multi-head attention which is useful to stabilise the learning \cite{Velickovic2018, Vaswani2017} by taking the expectation, $\mathbf{A}(\mathbf{X}) = \frac{1}{h}\sum_h \mathbf{A}^{h}(\mathbf{X}) $.
%
%
%
The attention weight matrix $\mathbf{A} = (a(\vec{X}_i,\vec{X}_j))$ is right-stochastic, allowing equation~\eqref{eq:graph_diffusion_node1} to be written as \vspace{-2mm}
\begin{equation}
\frac{\partial}{\partial t}\mathbf{X} = (\mathbf{A}(\mathbf{X})-\mathbf{I})\mathbf{X} = \bar{\mathbf{A}}(\mathbf{X})\mathbf{X}\vspace{-3mm}
    \label{eq:laplacian_form}
\end{equation}



%

As discussed in Section~\ref{sec:graph_diffusion}, a broad range of discretisations are possible. Temporal discretisations amount to the choice of numerical scheme, which can use either fixed 
or adaptive step sizes and be either explicit or implicit. 
Time forms a continuous analogy to the layer index, where each \textit{layer} corresponds to an iteration of the solver. When using adaptive time step solvers, the number of layers is not specified a-priori. 
Explicit schemes use residual structures (e.g. Figure\ref{fig:rk4_step}, left and middle) that are usually more complex than those employed in resnets and which follow directly from rigorous numerical stability results (see Appendix~\ref{sec:multistep}). 
Implicit numerical schemes offer a natural way of trading off \textit{depth} and \textit{width} (spatial support of the diffusion kernel). In Section~\ref{sec:discretisation_scheme} we explore several temporal discretisations using various numerical integrators.

Spatial discretisation amounts to modifying the given graph, or building one in settings where no graph is given and the data can be assumed to lie in some feature space or on a continuous manifold. When the input graph is given, which is the case in our experimental sections, we can rewire the given graph and use a different edge set in the diffusion equation.

While in general equation~(\ref{eq:laplacian_form}) is nonlinear due to the dependence of $\mathbf{A}$ on $\mathbf{X}$, it becomes linear if the attention weights are fixed inside the integral,  $\bar{\mathbf{A}}(\vec{X}(t)) = \bar{\mathbf{A}}$ (note that $\mathbf{A}$ is still parametric and learnable, but does not change throughout the diffusion process). 
In this case, equation~\eqref{eq:laplacian_form} can be solved analytically as $\vec{X}(t) = e^{\bar{\mathbf{A}}t} \vec{X}(0)$.
As $\bar{\mathbf{A}}$ is a form of normalised Laplacian, all eigenvalues are non-positive and the steady state solution is given by 
the dominating eigenvector, which is the degree vector. However, as $\bar{\mathbf{A}}$ is learned, this limitation is not severe as the system can be (and in practice is) degenerate; the graph becomes (approximately) disconnected, with connected components permitted to have unique steady state solutions.  We call this model \textbf{GRAND-l} for linear to distinguish it from the more general \textbf{GRAND-nl} for non-linear.
The final variant is \textbf{GRAND-nl-rw} (non-linear with rewiring), where rewiring is  performed via a two step process: as a preprocessing step, the graph is densified using diffusion weights as in~\cite{Klicpera2019}, and then at runtime the subset of edges to use is learned based on attention weights. 
Equation~\eqref{eq:graph_diffusion} becomes:
\begin{align}\vspace{-4pt}
 \frac{\partial\mathbf{X}_i(t)}{\partial t}  &= \sum_{j : (i,j) \in \mathcal{E}'}a\left(\vec{X}_i(t), \vec{X}_j(t)\right)\left(\vec{X}_j(t) - \vec{X}_i(t)\right)
 \label{eq:graph_diffusion_node1}
\vspace{-4pt}
\end{align}
where $\mathcal{E}' = \{(i,j):(i,j)\in \mathcal{E} \,\, \mathrm{and} \,\, a_{ij} > \rho\}$ with some threshold value $\rho$,  is the `rewired' edge set, which may now contain self-loops. While $a$ changes throughout the diffusion process, rewiring is only performed at the start of the epoch based on features at $t=0$.

GRAND shares parameters across layer/iteration and is thus more data-efficient than conventional GNNs. The full training objective is given in Appendix~\ref{sec:training_objective}. To update the parameters we either backpropagate through the computational graph of the numerical integrator or, when memory is constrained, use Pontryagin's maximum principle~\cite{pontryagin2018mathematical}.

\section{Related work}

\paragraph{Image processing and graphics.} 
During the 1990s-2000s, a vast amount of image processing literature exploited the formalism of diffusion equations \cite{weickert1998anisotropic}, starting with the seminal work of~\citet{perona1990scale}. 
\citet{sochen1998general} developed a differential geometric framework (`Beltrami flow') considering the evolution of images represented as embedded manifolds. 
The related bilateral \cite{tomasi1998bilateral} and non-local means \cite{buades2005non} filters, together with efficient numerical techniques \cite{Weickert1997,durand2002fast}, have popularised these ideas in the image processing community. 
PDE-based methods were also used for low-level tasks such as image 
segmentation \cite{caselles1997geodesic,chan2001active} and inpainting \cite{bertalmio2000image}. 

In computer graphics, solutions of non-Euclidean diffusion equations were studied as {\em heat kernel signature}  \cite{sun2009concise,bronstein2010scale} local shape descriptors related to the Gaussian curvature. 
Non-Euclidean diffusion equations can be solved by using the Laplacian eigenvectors as the analogy of Fourier basis and the corresponding eigenvalues as frequencies. The solution can be represented as a spectral transfer function \cite{patane2016star}, which can also be learned \cite{litman2013learning}. 
%
The non-Euclidean Fourier approach was exploited in the early work on deep learning on graphs  \cite{HeBrLe2015,defferrard2016convolutional,kipf2017,LeMoBrBr2017}.

\paragraph{Graph diffusion processes} 
techniques such as eigenmaps and diffusion maps \cite{coifman2005geometric, belkin2003laplacian} use linear diffusion PDEs with closed form solutions expressed through Laplacian eigenvectors. Diffusion-Convolutional Neural Networks \cite{atwood2016diffusion} employ a diffusion operator for graph convolutions and LanczosNet~\cite{LiaoZUZ19} uses a polynomial filter on the Laplacian matrix, which corresponds to a multi-scale linear diffusion PDE. Adaptive Lanczos-Net~\cite{LiaoZUZ19} additionally allows learning the filters to reweight the graph using a kernel. The use of a polynomial filter approximates the solution of the PDE, and the diffusion is linear with a fixed operator.
 
\paragraph{Neural ODEs.} 
\citet{Chen2018a} introduced neural ODEs. Many follow-up works  explored augmentation~\cite{Dupont2019} and  regularization~\cite{Finlay2020} and provided extensions into new domains such as stochastic ~\cite{sdes2020, tzen2019neural} differential equations.
Neural ODEs have also been applied to GNNs: 
\citet{Avelar2019} model continuous residual layers with GCN. \citet{Poli2019} propose approaches for static and dynamic graphs using GCN to model static graphs and a hybrid approach where the latent state evolves continuously between RNN steps for dynamic graphs.
\citet{Xhonneux2020} address continuous message passing. Their model is a solution to the constant linear diffusion PDE. Unlike most GNN, it scales with the size of the graph having $\mathcal{O}(n)$ parameters. Continuous GNNs were also explored by~\citet{gu2020implicit} who, similarly to~\cite{Scarselli2008}, addressed the solutions of fixed point equations. Ordinary Differential Equations on Graph Networks (GODE)\cite{Zhuang2020a} approach the problem using the technique of invertible ResNets. Finally, \citet{Sanchez-Gonzalez2019} used graph-based ODEs to generate physics simulations. 

\paragraph{Neural PDEs.}
Using deep learning to solve PDEs was explored by~\citet{Raissi2017a}. Neural networks appeared in~\cite{Li2020c} to accelerate PDE solvers with applications in the physical sciences. These have been applied to problems where the PDE can be described on a graph \cite{Li2020d}. \citet{belbute2020combining} consider the problem of predicting fluid flow and use a PDE inside a GNN. These approaches differ from ours in that they solve a given PDE, whereas we use the notion of discretising PDEs as a principle to understand and design GNNs. 

\section{Results}

We design experiments to answer the following: Are GNNs derived from the diffusion PDE competitive with existing popular methods? Can we address the problem of building deep graph neural networks? Under which conditions can implicit methods yield more efficient GNNs than explicit methods? Additional implementation details are provided in the Appendix. 

GRAND is implemented in PyTorch~\cite{torch}, using PyTorch geometric~\cite{torch_geo} and torchdiffeq~\cite{Chen2018a}. Code and instructions to reproduce the experiments are available at \url{https://github.com/twitter-research/graph-neural-pde}.

\subsection{Node classification benchmarks}

We measure the performance of GRAND on a range of common node classification 
benchmarks.

\paragraph{Methods} We compare to four of the most popular GNN architectures: Graph Convolutional Network (GCN)~\cite{kipf2017}, Graph Attention Network (GAT)~\cite{Velickovic2018}, Mixture Model Networks~\cite{Monti2016} and GraphSage~\cite{Hamilton2017}. 
Additionally we compare to recent ODE-based GNN models, Continuous Graph Neural Networks (CGNN)~\cite{Xhonneux2020}, Graph Neural Ordinary Differential Equations (GDE)~\cite{Poli2019}, and Ordinary Differential Equations on Graphs (GODE)~\cite{Zhuang2020a} and two versions of LanczosNet \cite{LiaoZUZ19} which approximate solutions to a linear diffusion PDE.

We study three variants of GRAND: linear, nonlinear and nonlinear with graph rewiring. In the GRAND-l, the attention weights are constant throughout the integration, producing a coupled system of linear ODEs. In GRAND-nl, the attention weights are updated at each step of the numerical integration. In both cases, the given graph is used as the spatial discretisation of the diffusion operator. In GRAND-nl-rw, the graph is rewired after each backward pass by thresholding the diffusivity attention mechanism. The rewiring is held constant throughout the integration.

\paragraph{Datasets}
We report results for the most widely used citation networks Cora~\cite{mccallum2000automating}, Citeseer~\cite{sen2008collective}, Pubmed~\cite{namata2012query}. These datasets contain fixed splits that are often used, which we include for direct comparison in Table~\ref{tab:planetoid_results}. To address the limitations of this evaluation methodology \cite{shchur2018pitfalls}, we also report results for all datasets using 100 random splits with 20 random initializations. Additional datasets are the coauthor graph CoauthorCS~\cite{shchur2018pitfalls}, the Amazon co-purchasing graphs Computer and Photo~\cite{mcauley2015image}, and the OGB arxiv dataset~\cite{hu2020open}. In all cases, we use the largest connected component. Dataset statistics are included in Appendix~\ref{ap:datasets}.

\paragraph{Experimental setup}
We follow the experimental methodology described in~\cite{shchur2018pitfalls} using 20 random weight initializations for datasets with fixed Planetoid splits and 100 random splits for the remaining datasets. 
Where available, results from~\cite{shchur2018pitfalls} were used. 
Hyperparameters with the highest validation accuracy were chosen and results are reported on a test set that is used only once. Hyperparameter search used Ray Tune~\cite{Liaw2018} with a thousand random trials using an asynchronous hyperband scheduler with a grace period of ten epochs and a half life of ten epochs. 
The code to reproduce our results is included with the submission and will be released publicly following the review process.
Experiments ran on  AWS p2.8xlarge machines, each with 8 Tesla V100-SXM2 GPUs.


\paragraph{Implementation details}

For smaller datasets (Cora, Citeseer) we used the Anode augmentation scheme~\cite{Dupont2019} to stabilise training. The  ogb-arxiv dataset used the Runge-Kutta method, for all others Dormand-Prince was used. 
%
For the larger datasets, we used kinetic energy and Jacobian regularization \cite{Finlay2020, Kelly2020}. 
The regularization ensures the learned dynamics is well-conditioned and easily solvable by a numeric solver, which reduced training time. 
We use constant initialization for the attention weights, $\mathbf{W}_K,\mathbf{W}_Q$, 
so training starts from a well-conditioned system that induces small regularization penalty terms \cite{Finlay2020}.

\paragraph{Complexity}

For all datasets we use the adjoint method described in~\cite{Chen2018a}. The space complexity is dominated by evaluating Equation~\eqref{eq:attention} over edges and is $\mathcal{O}(|\mathcal{E}'|d)$ where $\mathcal{E}'$ is the edge set following rewiring and $d$ is dimension of features. 
The runtime complexity is $\mathcal{O}(|\mathcal{E}'|d)(E_b + E_f)$, split between the forward and backward pass and can be dominated by either depending on the number of function evaluations ($E_b$, $E_f$).

\paragraph{Number of parameters}

In traditional GNNs there is a linear relationship between the number of parameters and depth. Conversely, GRAND {\em shares parameters across layers} (due to our choice of a time-independent attention) and consequently, requires significantly less parameters than competing methods, while achieving on par or superior performance. The versions of GCN, SAGE and GAT used for the ogb-arxiv results required 143K, 219K and 1.63M parameters respectively, while our model only 70K.  

\paragraph{Performance}
Tables~\ref{tab:planetoid_results}--\ref{tab:random_splits} summarise the results of our experiments. GRAND variants consistently perform among the best methods, achieving first place on all but one dataset, where it is second. 
On ogb-arxiv, our results are slightly inferior to the best-performing GAT, which, however, requires 20 times as many parameters. 







\begin{table}[!h]
\resizebox{0.5\textwidth}{!}{
\begin{tabular}{llll}
Planetoid splits & \textbf{CORA}       & \textbf{CiteSeer}   & \textbf{PubMed}     \\
\textbf{GCN}             & 81.9 $\pm$ 0.8 & 69.5 $\pm$ 0.9 & 79.0 $\pm$ 0.5 \\
\textbf{GAT}             & 82.8 $\pm$ 0.5 & 71.0 $\pm$ 0.6 & 77.0 $\pm$ 1.3 \\
\textbf{MoNet}           & 82.2 $\pm$ 0.7 & 70.0 $\pm$ 0.6 & 77.7 $\pm$ 0.6 \\
\textbf{GS-maxpool}      & 77.4 $\pm$ 1.0 & 67.0 $\pm$ 1.0 & 76.6 $\pm$ 0.8 \\
\textbf{Lanczos}    & 79.5$\pm$1.8 & 66.2$\pm$1.9 & 78.3$\pm$0.3 \\
\textbf{AdaLanczos}    & 80.4$\pm$1.1 & 68.7$\pm$1.0 & 78.1$\pm$0.4 \\
\textbf{CGNN$\dagger$}     & 81.7 $\pm$ 0.7 & 68.1 $\pm$ 1.2 & ${\bf 80.2 \pm 0.3}$ \\
\textbf{GDE*}    & $\textcolor{blue}{\bm{83.8 \pm 0.5}}$ & ${\bf 72.5 \pm 0.5}$ & 79.9 $\pm$ 0.3 \\
\textbf{GODE*}    & 83.3 $\pm$ 0.3 & 72.4 $\pm$ 0.6 & 80.1 $\pm$ 0.3 \\
\hline
\textbf{GRAND-l} (ours)          & $\textcolor{red}{\bm{84.7 \pm 0.6}}$ & $\textcolor{blue}{\bm{73.3 \pm 0.4}}$ & $\textcolor{blue}{\bm{80.4 \pm  0.4}}$  \\
\textbf{GRAND-nl} (ours)          & ${\bf 83.6 \pm 0.5}$ & $70.8 \pm 1.1$ & $79.7 \pm 0.3$ \\
\textbf{GRAND-nl-rw} (ours)          & 82.9 $\pm$ 0.7 & $\textcolor{red}{\bm{73.6 \pm 0.3}}$ & $\textcolor{red}{\bm{81.0 \pm 0.4}}$
\end{tabular}
}
\caption{Test accuracy and std for 20 random initializations using the original Planetoid train-val-test splits. *GODE and GDE comprises six and three separate models respectively. For each dataset we present the best performing variant of GODE and GDE. $\dagger$Results obtained running the authors' code with the hyperparameters given in their paper using Pytorch Geometric data readers.}
\label{tab:planetoid_results}
\end{table}

\begin{table*}[!h]
\resizebox{\textwidth}{!}{
\begin{tabular}{lccccccc}
Random splits & \textbf{CORA}       & \textbf{CiteSeer}   & \textbf{PubMed}  & \textbf{Coathor CS}       & \textbf{Computer}   & \textbf{Photo} & \textbf{ogb-arxiv$^*$}   \\
\textbf{GCN}           & 81.5 $\pm$ 1.3 & ${\bf 71.9 \pm 1.9}$ & 77.8 $\pm$ 2.9 & 91.1 $\pm$ 0.5 & 82.6 $\pm$ 2.4 & 91.2 $\pm$ 1.2  & ${\bf 72.17 \pm 0.33}$  \\
\textbf{GAT}            & 81.8 $\pm$ 1.3 & 71.4 $\pm$ 1.9 & $\textcolor{blue}{\bf 78.7 \pm 2.3}$ & 90.5 $\pm$ 0.6 & 78.0 $\pm$ 19.0 & 85.7 $\pm$ 20.3  & $\textcolor{red}{\bm{73.65 \pm 0.11}}^\dagger$ \\
\textbf{GAT-ppr}        & $81.6 \pm 0.3$ & $68.5 \pm 0.2$ & $76.7 \pm 0.3$  & $91.3 \pm 0.1$ & $\textcolor{blue}{\bm{85.4 \pm 0.3}}$ & $90.9 \pm 0.3$ & N/A  \\
\textbf{MoNet}           & 81.3 $\pm$ 1.3 & 71.2 $\pm$ 2.0 & ${\bf 78.6 \pm 2.3}$  & 90.8 $\pm$ 0.6 & 83.5 $\pm$ 2.2 & 91.2 $\pm$ 2.3 & N/A \\
\textbf{GS-mean}      & 79.2 $\pm$ 7.7 & 71.6 $\pm$ 1.9 & 77.4 $\pm$ 2.2 & 91.3 $\pm$ 2.8 & 82.4 $\pm$ 1.8 & 91.4 $\pm$ 1.3 & 71.39 $\pm$ 0.16 \\
\textbf{GS-maxpool}      & 76.6 $\pm$ 1.9 & 67.5 $\pm$ 2.3 & 76.1 $\pm$ 2.3 & 85.0 $\pm$ 1.1 & N/A & 90.4 $\pm$ 1.3 &  N/A \\
\textbf{CGNN}     & $81.4 \pm 1.6$ & $66.9 \pm 1.8$ & 66.6 $\pm$ 4.4 & ${\bf 92.3 \pm 0.2}$ & 80.29 $\pm 2.0$ & 91.39 $\pm$ 1.5 & 58.70 $\pm$ 2.5 \\
\textbf{GDE} & 78.7 $\pm$ 2.2 & 71.8 $\pm$ 1.1 & 73.9 $\pm$ 3.7 & 91.6 $\pm$ 0.1 & 82.9 $\pm$ 0.6 & $\textcolor{blue}{\bm{92.4 \pm 2.0}}$ & 56.66 $\pm$ 10.9 \\
\hline
\textbf{GRAND-l} (ours)          & $\textcolor{red}{\bm{83.6 \pm 1.0}}$   & $\textcolor{blue}{\bm{73.4 \pm 0.5}}$ & $\textcolor{red}{\bm{78.8 \pm  1.7}}$ & $\textcolor{red}{\bm{92.9 \pm 0.4}}$ & ${\bf 83.7 \pm 1.2}$ &  $\bm{92.3 \pm 0.9}$ & 71.87 $\pm$ 0.17 \\
\textbf{GRAND-nl} (ours)          & ${\bf 82.3 \pm 1.6}$ & $70.9 \pm 1.0$ & $77.5 \pm 1.8$
& $\textcolor{blue}{\bm{92.4 \pm 0.3}}$ &  $82.4 \pm 2.1$ & $\textcolor{blue}{\bm{92.4 \pm 0.8}}$ & 71.2 $\pm$ 0.2  \\
\textbf{GRAND-nl-rw} (ours)          & $\textcolor{blue}{\bm{83.3 \pm 1.3}}$ & $\textcolor{red}{\bm{74.1 \pm 1.7}}$ & $78.1 \pm 2.1$ & $91.3 \pm 0.7$ &  $\textcolor{red}{\bm{85.8 \pm 1.5}}$ & $\textcolor{red}{\bm{92.5 \pm 1.0}}$ &  $\textcolor{blue}{\bm{72.23 \pm 0.20}}$ \\
\end{tabular}
}\vspace{-2mm}
\caption{Test accuracy and std for 20 random initializations and 100 random train-val-test splits. *Using labels. $^\dagger$using 1.5M parameters. }
\label{tab:random_splits}
\end{table*}

\subsection{Depth}
\begin{figure}[!]
    \centering\vspace{-4mm}
    \includegraphics[width=\columnwidth]{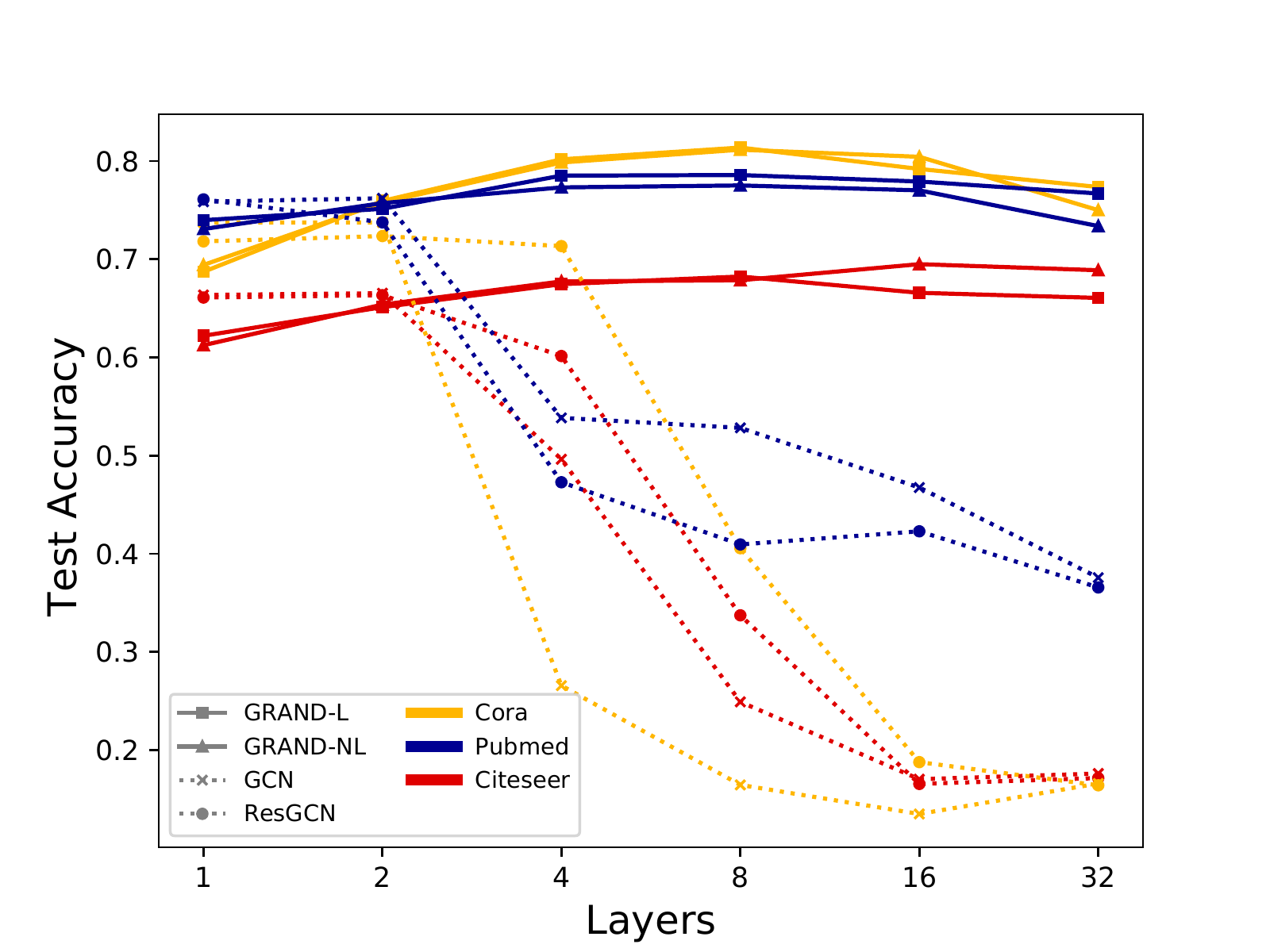}\vspace{-3mm}
    \caption{Performance of architectures of different depth. }\vspace{-3mm}
    \label{fig:depth}
\vspace{-3mm}
\end{figure}

To demonstrate that our model solves the oversmoothing problem and performs well with many layers, we performed an experiment using the RK4 fixed step-size solver (with step size $\tau=1.0$), varying the integration time $T$ while holding the other hyper-parameters fixed. This effectively produces architectures of varying depth. Figure \ref{fig:depth} shows that compared to GCN and a GCN with residual connections,  our model maintains performance as the layers increase whilst the baselines degrade by $50\%$ after 4 layers. 

\subsection{Choice of discretisation scheme}
\label{sec:discretisation_scheme}

\begin{figure*}[]
    \centering
    \begin{subfigure}{\columnwidth}
    \includegraphics[width=\columnwidth]{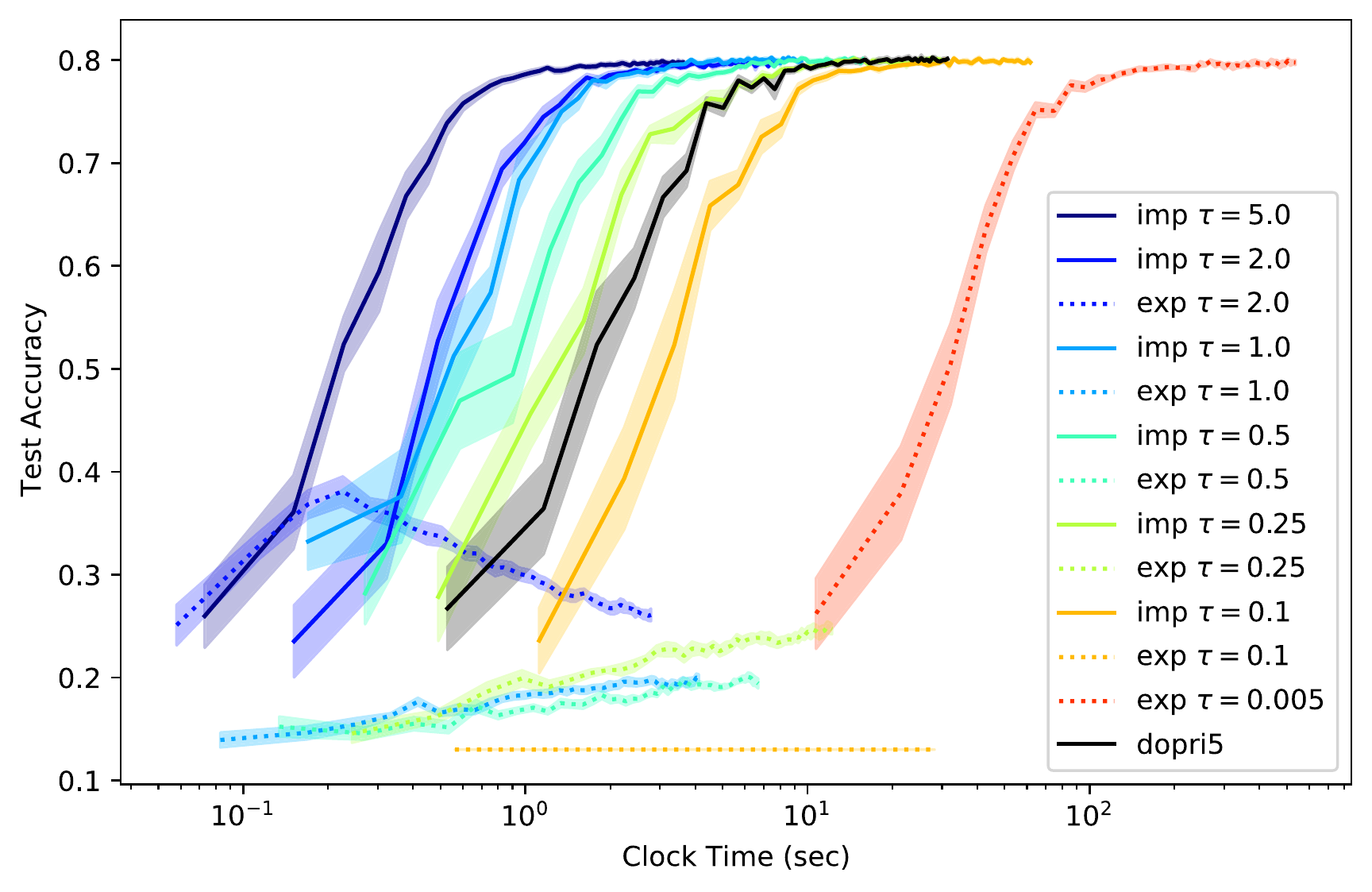}
    \label{fig:implicit-vary-stepsizes}
    \end{subfigure}
    \begin{subfigure}{\columnwidth}
        \includegraphics[width=\columnwidth]{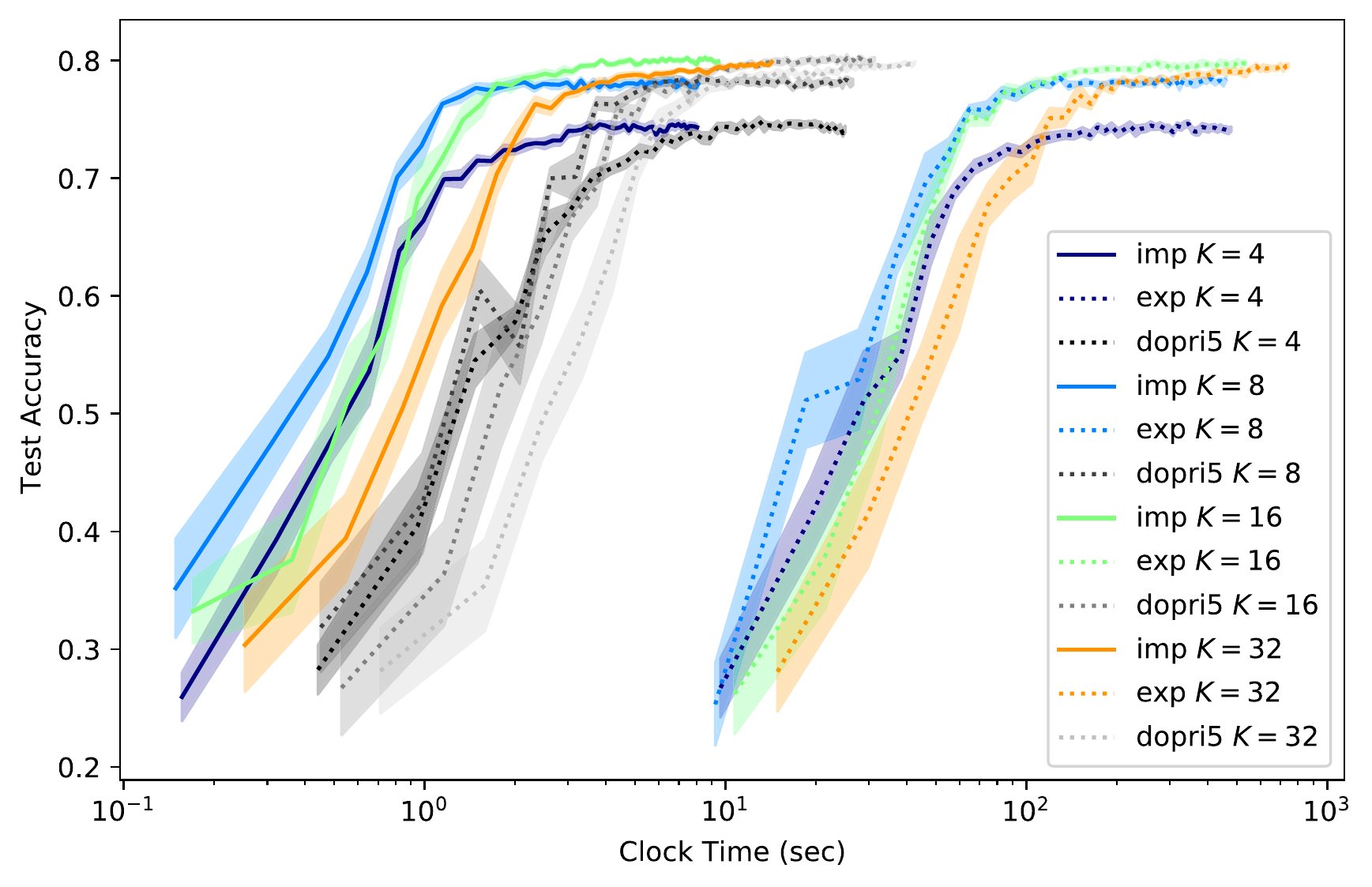}
    \label{fig:implicit-vary-K}
    \end{subfigure}\vspace{-6mm}
    \caption{Performance of different solvers. Left: Test accuracy on plain Cora varying the step size, comparing explicit Adams--Bashford, implicit Adams--Moulton, and adaptive Runge-Kutta 4(5). We observe that the explicit fixed step size scheme is unstable for all but a small step size whereas the implicit scheme is stable for all step sizes tried. For a large step size, the implicit scheme is faster than a state-of-the-art explicit scheme with adaptive step size. Right: Diffusion-rewired Cora varying the sparsity of the graph by keeping the largest $K$ coefficients for each node. Explicit Adams--Bashford has step size $\tau=0.005$, and implicit Adams--Moulton $\tau=1.0$. We observe a trade-off between sparsity, speed, and accuracy.}\vspace{-3mm}
    \label{fig:implicit-exps}
\end{figure*}

We investigated the stability of explicit numerical schemes with a fixed step size and the tradeoff between step size and computational time for an equivalent implicit numerical scheme. We compared these to the Dormand–Prince adaptive step size scheme (DOPRI5).

\paragraph{Method choice}
We ran GRAND on Cora with the explicit Adams--Bashford method, an implicit Adams--Moulton method with a predictor-corrector algorithm, and the adaptive Runge-Kutta 4(5) method (see Figure \ref{fig:implicit-exps}, left), varying the step sizes for the two fixed-step size methods. We observe that the explicit Adams method is unstable for all but a small step size of $\tau=0.005$, while the implicit Adams method is stable for all step sizes. Moreover, in this case the implicit method converges to the solution faster than a state-of-the-art adaptive step size solver for large enough step size. We note, however, that this may not always be the case. As the step size is increased, the implicit method can take fewer steps. However, as the step size is increased the implicit equations become more difficult to solve and require more iterations of the algorithm used to solve them.

\paragraph{Graph rewiring}
In this experiment, we rewired the Cora graph using the method of \citet{Klicpera2019}, keeping the largest $K$ coefficients for each node. We varied $K$ to explore the tradeoff between sparsity, computation time, and accuracy (see Figure~\ref{fig:implicit-exps}, right). As the graph is made sparser ($K$ decreases), all methods become faster. 
The accuracy converges to similar values until the graph is so sparse that the flow of information is impeded ($K<8$).
We observe that for implicit solvers the benefit of sparsification is independent of the step size, and both can be combined somewhat to decrease the time per epoch without effecting accuracy. We hypothesize that, in general, a sparser graph is particularly desirable for implicit solvers since it may reduce the difficulty of solving the implicit equations (less iterations until convergence).
A final observation is that we can draw a  diagram similar to Figure \ref{fig:rk4_step} for the computational graph of the Adams--Moulton method with predictor--corrector steps; it is redolent of an RNN with adaptive computation time \citep{DBLP:journals/corr/Graves16}, where the stopping rule is deterministic rather than learnt (that is,  continue unrolling the RNN until the difference between outputs is below a cutoff).

%
%

\subsection{Diffusion on MNIST Image Data Experiments}

\begin{figure}
    \vspace{-1mm}
    \centering
    \includegraphics[width=\columnwidth]{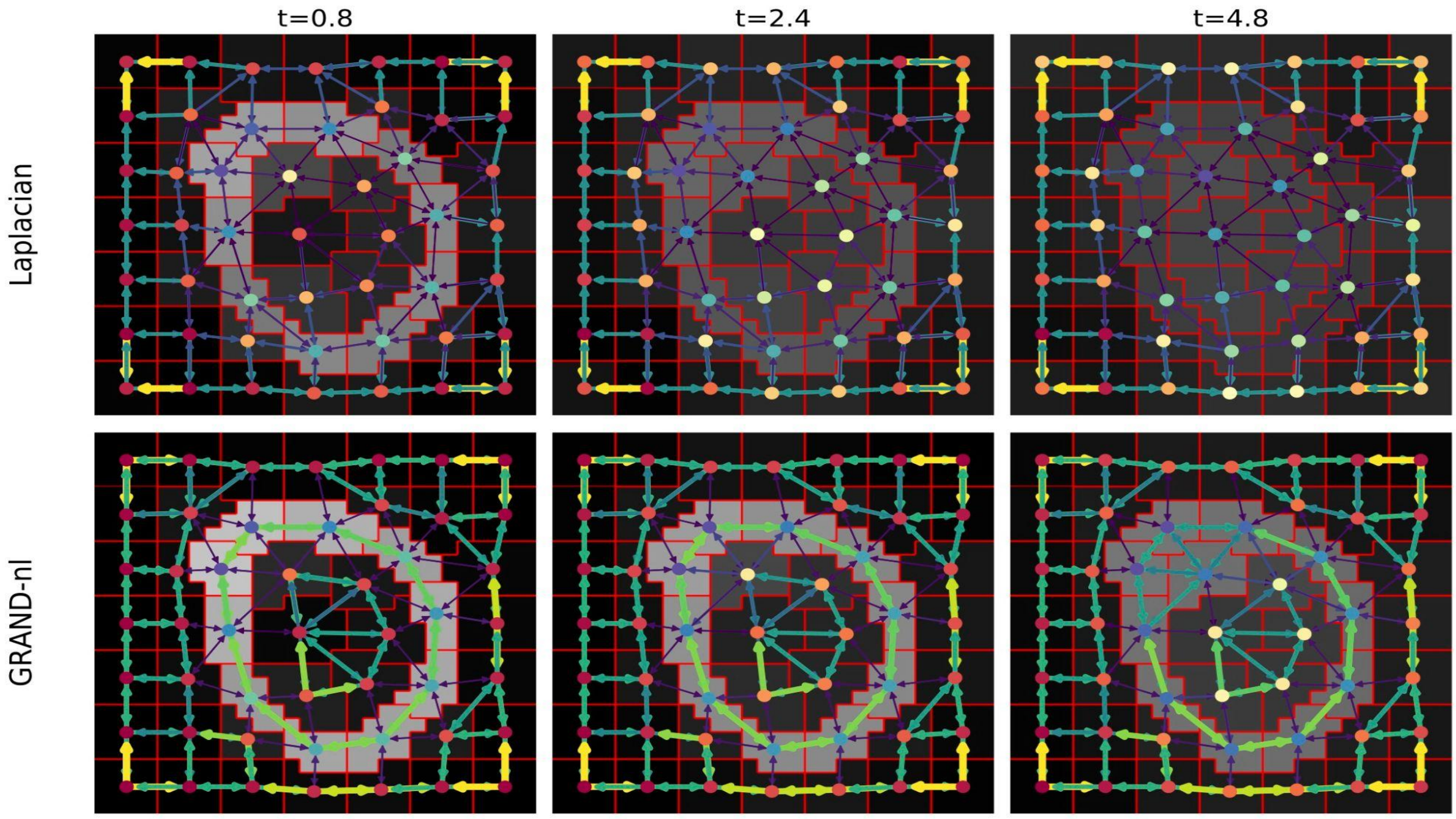}
    \vspace{-8mm}
    \caption{Illustration of the effect of attention weights on pixel diffusion}
    \label{fig:MNIST_diff}
    \vspace{-7mm}
\end{figure}

We performed an experiment to illustrate the learned diffusion characteristics of GRAND. MNIST pixel data was used to construct a superpixel representation \cite{slic} and adjacent patches were joined with edges, binary pixel labels were applied (number or background) with a 50\% training mask. We evolved both GRAND-nl and a constant Laplacian diffusion model for $T=4.8$ and $\tau=0.8$, equating to a 6 layer GNN. We show the attention weights by the colour and thickness of the edges. Figure~\ref{fig:MNIST_diff} shows Non-linear GRAND performs edge detection weighting diffusion within a class boundary in a way that preserves the image after diffusion. The Laplacian diffusion is unable to preserve the features of the original image.

\section{Conclusion}

We presented a new class of graph neural network called Graph Neural Diffusion (GRAND), based on the discretisation of diffusion PDEs on graphs. 
Our framework allows leveraging vast literature on PDEs relating to discrete temporal and spatial operators and stability, and provides a blueprint for a principled design of new graph learning architectures. 
We show that appropriate choice of discretisation and numerical schemes in GRAND allows us to train very deep graph neural networks and results in superior performance on popular benchmarks. 

\paragraph{Limitations}
We intentionally considered a form of the diffusion equation that is easier to treat mathematically. 
Our model is currently limited to learn only functions of the form $\frac{\partial\vec{x}}{\partial t} = f(\vec{x}(t), t, \theta)$, with an `attentional' structure of $f$. This imposes two limitations that are not present in discrete neural networks: first, the size of the hidden state vector must be constant for all layers (a usual situation in GNNs), and second, the same set of parameters $\theta$ must be used for all layers. The later constraint comes as an advantage, allowing our model to use $10- 20$ times less parameters than the top performing model on ogbn-arxiv. In future work, we intend to overcome these limitation by introducing a $\theta = \theta(t)$ as described in~\cite{Queiruga2020, Zhang2019c}.
%
We will also consider more general nonlinear diffusion equations that result in message passing `flavors' of GNN architectures. 

\paragraph{Acknowledgements} MB is supported in part by ERC Consolidator grant No. 724228 (LEMAN). We would like to thank Gabriele Corso, Nils Hammerla and our reviewers for many helpful suggestions that improved this manuscript.

\appendix

\section{Datasets} \label{ap:datasets}

The statistics for the largest connected components of the experimental datasets are given in Table~\ref{tab:dataset_stats}.

\begin{table*}[!h]
\begin{tabular}{ccccccc}
\centering
\textbf{Dataset} & \textbf{Type} & \textbf{Classes} & \textbf{Features} & \textbf{Nodes} & \textbf{Edges} & \textbf{Label rate} \\
\hline
Cora             & citation      & 7                & 1433              & 2485           & 5069           & 0.056               \\
Citeseer         & citation      & 6                & 3703              & 2120           & 3679           & 0.057               \\
PubMed           & citation      & 3                & 500               & 19717          & 44324          & 0.003               \\
Coauthor CS      & co-author     & 15               & 6805              & 18333          & 81894          & 0.016               \\
Computers        & co-purchase   & 10               & 767               & 13381          & 245778         & 0.015               \\
Photos           & co-purchase   & 8                & 745               & 7487           & 119043         & 0.021   \\        
OGB-Arxiv & citation & 40 & 128 & 169343 & 1166243 & 1
\end{tabular}
\caption{Dataset Statistics}
\label{tab:dataset_stats}
\end{table*}

\section{Diffusivity Formulations} \label{ap:diffusivity}

GRAND can use any right stochastic attention matrix. We performed experiments with the multiheaded Bahdanau formulation~\cite{Bahdanau2014} of attention, which has previously been applied to graphs in~\cite{Velickovic2018}
\begin{align}
a(x_i,x_j) = \frac{\exp \left(\leakyrelu \left(\mathbf{a}^{T}\left[W \vec{x}_{i} \| W \vec{x}_{j}\right]\right)\right)}{\sum_{k \in \mathcal{N}_{i}} \exp \left(\leakyrelu\left(\mathbf{a}^{T}\left[W \vec{x}_{i} \| W \vec{x}_{k}\right]\right)\right)},
\label{eq:bahdanau_attention}
\end{align}
where $W$ and $\vec{a}$ are learned and $\|$ is the concatenation operator. However, for all datasets, the scaled dot product attention performed better. 
This may be because GAT relies on dropout. Dropout performs poorly inside adaptive timestep numerical ODE solvers as the stochasticity in the forward pass drives $\tau \to 0$.

\section{Full Training Objective} \label{sec:training_objective}

The full training program optimises cross entropy loss
\begin{align}
    \mathcal{L}(\mathbf{Y},\mathbf{T}) &= H(\mathbf{Y},\mathbf{T}) =  \sum_{i=1}^{n} \vec{t}^T_i \log \vec{y}_i 
\end{align}
where $\vec{t}_i \in \mathbb{R}^{d_{\text{class}}}$ is the one-hot truth vector of the $i^{th}$ node with prediction
\begin{align}
    \vec{y}_i &= \psi(\mathbf{x}_i(T)) 
\end{align}

where $\psi: \mathbb{R}^{d}\rightarrow \mathbb{R}^{d_{\text{class}}}$ is a linear layer decoder of the terminal value of the evolutionary PDE
\begin{align}
  \vec{y}_i  &= D\mathbf{x}_i(T) + \vec{b}_d \\
  &= D\left(\mathbf{X}(0) + \int_0^T \frac{\partial\mathbf{X}(t)}{\partial t} dt\right) + \vec{b}_d \\
  &= D\left(\phi(\mathbf{X}_{\mathrm{in}}) + \int_0^T \frac{\partial\mathbf{X}(t)}{\partial t} dt \right) + \vec{b}_d 
\end{align}
with the initial condition given by the linear layer encoder $\phi:\mathbb{R}^{d_{\text{in}}}\rightarrow \mathbb{R}^{d}$ of the input data
\begin{align*}
    \vec{y}_i  &= \left(D\left(E(\mathbf{X}_{\mathrm{in}})+\vec{b}_e + \int_0^T \frac{\partial\mathbf{X}(t)}{\partial t} dt \right) + \vec{b}_d\right)_{i} 
\end{align*}
and $f(\vec{X}(t), t, \theta) = \frac{\partial\mathbf{X}(t)}{\partial t}$ is the system dynamics that we wish to learn. For the nonlinear version of GRAND this is 
\begin{align}
f = \frac{\partial}{\partial t}\mathbf{X}(t) = (\mathbf{A}(\mathbf{X}(t))-\mathbf{I})\mathbf{X}(t) = \bar{\mathbf{A}}(\mathbf{X}(t))\mathbf{X}(t) \notag
\end{align}

\section{Stability} \label{ap:stability}
\subsection{Stability of linear ODE}

In the main paper we reported the linear GRAND   $\dot{\mathbf{x}}=\bar{A}\mathbf{x}$ has solution 
\begin{align}
    \vec{x}(t) = \vec{x}(0)e^{\bar{A}t}.
\end{align}
As $\bar{A}$ is not diagonal this matrix exponential is not analytically recoverable. Performing eigenvalue decomposition the solution is
\begin{align}
    \mathbf{x}(t)=\bar{T}e^{\bar{D}t}\bar{T}^{-1}\mathbf{x}(0).
\end{align}
Assuming $\bar{T}^{-1}$ exists, $\bar{T}$ has full rank and both are bounded, the test equation becomes
\begin{align}
    \mathbf{y}(t)=e^{\bar{D}t}\mathbf{y}(0)
\end{align} 
where $\mathbf{y}(t)=\bar{T}\mathbf{x}(t)$. If $\mathbf{x}(t)$ and $\mathbf{\hat{x}}(t)$ are two solutions of the ODE then their projections in eigenspace are $\mathbf{y}(t)$ and $\mathbf{\hat{y}}(t)$. For each node $i$:
\begin{align}
|y_i(t)-\hat{y}_i(t)|&=\left|(y_i(0)-\hat{y}_i(0)) e^{\bar{\lambda}_{i} t}\right|\\
&=|y_i(0)-\hat{y}_i(0)| e^{\mathcal{R}e(\bar{\lambda}_{i}) t}
\end{align}
for this to converge as $t \rightarrow \infty$ we require $\mathcal{R}e(\bar{\lambda}_{i})\leq0$ $\forall i$. As $A$ is right stochastic the eigenvalues of $\bar{A}=A-I$ satisfy this property.

\section{Numerical Schemes} \label{ap:schemes}
\subsection{Proof of theorem 1: Stability of explicit Euler}
For linear GRAND with an Euler numerical integrator 
\begin{align}
\vec{x}^{(t+1)} &= \left(I + \tau \bar{A}(\vec{x}^{(t)})\right)\vec{x}^{(t)} \\
&= Q^{(t)}\vec{x}^{(t)}.
\end{align}
We require that the amplification factor $||Q^{(t)}|| < 1$. It is sufficient to show that $Q^{(t)}$ is a right stochastic matrix, which has the property that its spectral radius $\lambda_{\max} \leq 1$. $Q$ is right stochastic if 
\begin{enumerate}
\item $\sum_{j=1}^{N} q_{ij} = 1$ 
\item $q_{ij} > 0 \quad \forall i,j$ 
\end{enumerate}
as $A$ is right stochastic $\sum_jI_{ij}+\tau(A_{ij}-I_{ij}) = 1$ proving 1). As $a_{ij}=q_{ij}$ for $i\neq j$, to prove 2) it remains to show that $1 + \tau (a_{ii} -1) > 0$ $\iff \tau < 1$.

\subsection{Proof of theorem 2: Implicit methods}
For implicit Euler
\begin{align}
    \dot{x}_n &= \frac{x_n - x_{n-1}}{\tau} = f(x_n, t_n) \\
    x_n &= \tau f(x_n, t_n) + x_{n-1},
\end{align}
incrementing the indices gives
\begin{align}
    x_{n+1} = \tau f(x_{n+1},t_{n+1}) + x_n 
\end{align}
and now, unlike the explicit case, $x_{n+1}$ now appears on both sides of the equation. If $f$ is linear
\begin{align}
    x_{n+1} &= \tau \bar{A} x_{n+1} + x_n \\
    x_{n+1} &= (I - \tau \bar{A})^{-1}x_n = B^{-1}x_n = Qx_n,
\end{align}
and the matrix $B$ must be inverted. The inverse exists as $B$ is diagonally dominant
\begin{align}
    I_{ii} - \tau(A_{ii} - I_{ii}) > \tau \sum_{j\neq i} A_{ij} = \tau(1 - A_{ii})
\end{align}
By considering the action of $B$ on $w=(1, ... , 1)^T$ it is clear that $Bw=w \implies Qw=w \implies \sum_j Q_{ij}=1$. As $B$ is diagonally dominant it is irreducible and satisfies $B_{ij} \leq 0 \,\, i \neq j$ and $B_{ii} > 0$ giving
$Q_{ij} > 0 \,\, \forall i,j$~\cite{varga1999matrix} and $Q$ is a Markov matrix with spectral radius bounded by unity and the implicit scheme is stable for all choices of $\tau$.

\section{General Multistep Methods} \label{sec:multistep}

A general multistep method (combining both implicit and explicit methods) can be written as
\begin{align}
    x_{n+1} + \sum_{i=1}^s \alpha_i x_{n+1-i} = \tau \sum_{i=0}^s \beta_if_{n+1-i},
\end{align}
where $f = \dot{x}$. If $\beta_0 = 0$ then $x_{n+1}$ only depends on terms up to $n$ and the method is explicit. 

\subsection{Order}

The order of a method gives the approximation error in terms of a Taylor series expansion.
If $p$ is the order, then the error is a single step $\propto \tau^{p+1}$ and the error in the entire interval $\propto \tau^p$.
In practice the order of a numerical method can be determined by measuring how the error changes with step size for a known integral.

\subsection{Butcher Tableau}

The set of coefficients for each multi step method are given by the Butcher Tableau. The simple case of forward Euler has $\alpha_1=-1$, $\beta_1 = 1$ with all other terms zero.

There is a law of diminishing return that relates the minimum number of function evaluations and the order of a higher order Runge-Kutta solver. Table~\ref{tab:order_nfe} shows why the Runge-Kutta 4 method (RK4) is often regarded as the optimal trade-off between speed and accuracy for multi step solvers.
\begin{table}[]
    \centering
    \begin{tabular}{c|cccccccc}
         Order & 1 & 2 & 3 & 4 & 5 & 6 & 7 & 8  \\
         Evals & 1 & 2 & 3 & 4 & 6 & 7 & 9 & 11
    \end{tabular}
    \caption{Function evaluations grow super-linearly with order after 4.}
    \label{tab:order_nfe}
\end{table}\textbf{}

\subsection{Runge-Kutta 4}

For all experiments we find that Runge-Kutta 4 (or it's adaptive step size variants) outperforms lower order methods. The Runge-Kutta 4 method follows the schema: if $f(\vec{x},t) = \bar{A}(\vec{x}_t)\vec{x}_t$
\begin{align}
    \vec{x}^{(t+1)} &= \vec{x}^{(t)} + \frac{1}{6}\tau \left(\vec{k}_1 + 2\vec{k}_2 + 2\vec{k}_3 + \vec{k}_4 \right) \\
    \vec{k}_1 &= f(\vec{x}_t, t) \\
    \vec{k}_2 &= f(\vec{x}_t + \tau \vec{k}_1/2, t+\tau / 2) \\
    \vec{k}_3 &= f(\vec{x}_t + \tau \vec{k}_2/2, t+\tau / 2) \\
    \vec{k}_4 &= f(\vec{x}_t + \tau \vec{k}_3, t + \tau)
\end{align}

\subsection{Adaptive Step Size}

Adaptive step size solvers estimate the error in $x_{n+1}$, which is compared to an error tolerance. The error is estimated by comparing two methods, one with order $p$ and one with order $p-1$. They are interwoven, i.e., they have common intermediate steps. As a result, estimating the error has little or negligible computational cost compared to a step with the higher-order method.

\begin{align}
x_{n+1}^{*}=x_{n}+\tau\sum _{i=1}^{s}b_{i}^{*}k_{i}
\end{align}
where $k_{i}$ are the same as for the higher-order method. Then the error is

\begin{align}
e_{n+1} = x_{n+1}-x_{n+1}^* = \tau\sum_{i=1}^s (b_{i}-b_i^*)k_i. 
\end{align}

The time step is increased if the error is below tolerance and decreased otherwise.

\section{Adaptive step size implementation details} \label{ap:architecture}


Most results presented used the adaptive step size solver Dormand-Prince5. Key to getting this to work well is setting appropriate tolerances for the step size.
Adaptive step size ODE solvers require two tolerance parameters; the relative tolerance rtol and the absolute atol. Both are used to assess the new step size
\begin{align}
    etol = atol + rtol * \max(|x_0|, |x_1|),
\end{align}
where $x_0$ and $x_1$ are successive estimations of the new state. 
\citet{Dupont2019} speculate that ResNets can learn a richer class of functions than ODEs because ``the error arising from discrete steps allows trajectories to cross''. 
We find that increasing the estimation error is also helpful when learning continuous diffusion functions and use value of $rtol$ and $etal$ that are $\times 10-\times 1000$ larger than the defaults. This both improves prediction accuracy and reduces the runtime.

In hyperparameter search $atol$ and $rtol$ were paired together using a tolerance scale variable $ts$ such that $atol = ts \times 10^{-12}$ and $atol = ts \time 10^{-6}$.

When using the adjoint method to backpropagate derivatives, two separate ODEs are being solved. This requires separate tolerance scales, which may differ: the forward pass tolerance, $ts$, controls for how close the approximated ODE solution is compared to the true solution, while the backward pass tolerance, $ts_{adj}$, controls the accuracy of the computed gradient. 
The hyperparameter search includes both $ts$ and $ts_{adj}$.

\clearpage

\bibliographystyle{icml2021}
\newcommand{\showDOI}[1]{\unskip} 
\newcommand{\showURL}[1]{\unskip} 
\bibliography{references}

\end{document}